\pdfoutput=1

\documentclass[11pt]{article}

\usepackage[]{acl}

\usepackage{times}
\usepackage{latexsym}
\usepackage{amsmath}
\usepackage{float}
\usepackage{algpseudocode}
\usepackage{algorithm}

\usepackage{enumitem}

\usepackage[T1]{fontenc}

\usepackage[utf8]{inputenc}

\usepackage{microtype}

\usepackage{graphicx}
\usepackage{todonotes}
\usepackage{adjustbox}

\title{\textit{What makes you change your mind?} \\ An empirical investigation in online group decision-making conversations}

\author{Georgi Karadzhov \\
  University of Cambridge \\
   \texttt{georgi.karadzhov@cl.cam.ac.uk}
  \\\AND
  Tom Stafford \\
  University of Sheffield \\
     \texttt{t.stafford@sheffield.ac.uk}
  \\\And
  Andreas Vlachos \\
  University of Cambridge\\ 
\texttt{av308@cam.ac.uk}
}
\begin{document}
\maketitle
\begin{abstract}
People leverage group discussions to collaborate in order to solve complex tasks, e.g.\ in project meetings or hiring panels. By doing so, they engage in a variety of conversational strategies where they try to convince each other of the best approach and ultimately reach a decision. In this work, we investigate methods for detecting \textit{what} makes someone change their mind.
To this end, we leverage a recently introduced dataset containing group discussions of people collaborating to solve a task.
To find out what makes someone change their mind, we incorporate various techniques such as neural text classification and language-agnostic change point detection. Evaluation of these methods shows that while the task is not trivial, the best way to approach it is using a language-aware model with learning-to-rank training. 
Finally, we examine the cues that the models develop as indicative of the cause of a change of mind.

\end{abstract}

\section{Introduction}
Research in group decision-making has shown that a group that collaborates in order to make a decision can outperform even the most knowledgeable individual within it \cite{mercier2011humans}. People engage in discussions in a variety of settings, such as project meetings and study groups. In these scenarios, people incorporate a variety of conversational strategies to introduce their ideas and convince each other of them, aiming ultimately to reach a consensus. Fundamentally, before committing to a decision, most of the participants in a group have different ideas of what the correct answer might be, but through discussion they are able to convince each other, and ultimately some of the participants change their mind. 
While previous research has shown that people who reach a consensus tend to perform better at certain tasks \citep{navajas2018aggregated, niculae2016conversational, concannon2015shifting}, \textit{how} people reach a consensus is understudied.  
Successfully identifying what makes someone change their mind, is an important step in studying group dynamics, persuasion and collaboration. 

\begin{figure}[h!]
    \centering
    \includegraphics[width=0.45\textwidth]{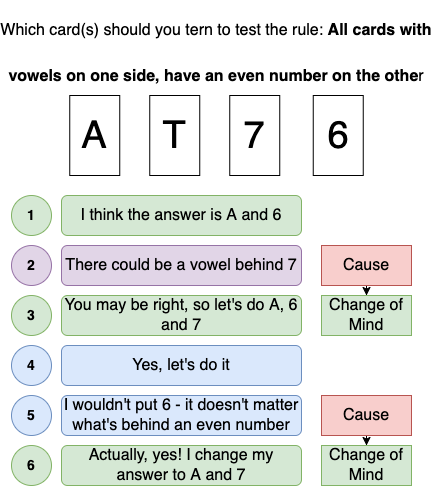}
    \caption{Sample conversation containing change of mind and what caused it. Participants are solving the Wason card selection task, where they should pick cards with letters and numbers on them.}
    \label{fig:example_conversation_intro}
\end{figure}

In this work, we take advantage of a dataset previously introduced by us \citep{karadzhov2021delidata}, which contains group discussions of people solving a cognitive task. The dataset contains 500 dialogues, where people engage in various deliberation patterns to communicate their solution to the problem.
The participants are presented with the Wason card selection task \cite{wason1968reasoning}, which is a classic problem used in the study of decision making and has been useful in testing the potential benefits and mechanisms of group discussion \citep{maciejovsky2013teams}. The Wason card selection task provides a controlled setup with quantifiable measures of success and improvement, which makes it very suitable for the study of individual biases and strategies. In the example in Figure~\ref{fig:example_conversation_intro}, participants engage in a collaborative discussion where they iterate through 3 different solutions, where one of the participants changes their mind twice (in utterances 3 and 6). In the example, conversation utterances 2 and 5 are the arguments that cause that change of mind, and are the target utterances that we would like to predict. Put formally, in order to investigate \textit{what} made someone change their mind in group decision-making conversations we 
formalise the task as detecting the utterance that causes the change of mind (or conversational turning point, which is used in this paper interchangeably).

In this work, we draw similarities between conversational turning points and change point detection. Change point detection investigates when a change will occur in a stream of data, and is traditionally applied in domains such as finance \citep{chen1997testing, oh2001intelligent}, engineering \citep{turner2013online, lai1995sequential}, climate data \citep{reeves2007review, khapalova2018assessing}, and genetics, \citep{wang2011non_cp_genes, hensman2013hierarchical}. Change point detection is concerned with either identifying a change post-hoc (offline change point detection or segmentation), or predicting a change point before it occurs - online change point detection \citep{adams2007bayesian}. In this work, we are concerned with the latter - identifying a change of mind before it occurs. We are doing this by trying to predict which utterance will cause a change of mind. 

\par

To evaluate our approaches, we develop a framework that quantifies the performance of models for change point detection in conversations, adopting practices from previous work \citep{burg2020evaluation}.
In terms of modelling, we first adapt a method for Bayesian online change point detection \cite{adams2007bayesian}, that was previously used in engineering and finance.
This method is language-agnostic as it ignores any kind of linguistic cues.
Next, we explore standard approaches to text classification as a method for predicting conversational turning points, showing that they are comparable to the language-agnostic models.
We further improve on these methods, by investigating learning-to-rank training for the prediction of what causes a change of mind. We demonstrate that by altering the training procedure and by incorporating the RankNet loss \citep{burges2005learning_ranknet}, we can substantially improve over the language-agnostic and text classification approaches.



Overall, our results demonstrate that the task of detecting conversational turning points is feasible but not trivial. Approaches such as bag-of-words, or a simple all-positive baseline for change point detection have a performance of 0.18 area under the precision-recall curve. On the other hand, a combination of our learning-to-rank model and a positional prior led to an AUC of 0.25. 
Finally, we conclude this study with a qualitative analysis, where we demonstrate different patterns and linguistic phenomena that may indicate a cause of change of mind.

\section{Related Work}

The effect of conversation in group decision-making
has previously been explored in the field of psychology. Both \citet{navajas2018aggregated} and \citet{mercier2011humans} show that there are conditions where a group of people who collaborate on a problem can outperform even the best individual group member. Moreover, previous research \citep{navajas2018aggregated, niculae2016conversational} has shown that groups of people who can reach a consensus through discussion have a higher group performance than just discussing or voting on a solution. 
\citet{concannon2015shifting} has found that disagreement markers at the beginning of the conversation lead to productive discussions. Likewise, both \citet{de2021beg} and \citet{hallsson2020disagreement} study the effect that disagreement has on constructive conversations, showing how people who are disagreeing with each other can work together.  Therefore, we hypothesise that it is interesting to study the conversations where someone changes their mind i.e.\ they disagreed at first but ultimately reach a consensus. 
\par
Other research is concerned with which specific linguistic phenomena are associated with conversations that can change someone's mind (or persuasive conversations). \citet{zeng2020whatchanged} investigate how topics and discourse change during a conversation, as well as their contribution to the persuasiveness of the conversation. Similarly, \citet{hidey-etal-2017-analyzing} analysed the prevalence of claims and premises in persuasive vs.\  non-persuasive dialogues. Both of these papers leverage the online forum \textit{Change my View}, where participants argue pro and against a certain topic. Unfortunately, the topics discussed in this forum are open to interpretation and personal opinion. Therefore, they do not have a clear quantitative measure of whether someone changed their mind. 
Further, while \citet{zeng2020whatchanged} and \citet{hidey-etal-2017-analyzing} study which phenomena may indicate that the conversation is more persuasive, they do not try to predict \textit{when} will someone change their mind, and what is the specific utterance that caused that.

Identifying when a change in a sequence of observations occurs is traditionally studied in the context of change point detection, in the field of signal processing \citep{page1954continuous, truong2020selective}.  Formally defined, if we observe a sequence of a variable $[x_1, x_2, ..., x_n]$, a change point occurs when two adjacent elements of that observation differ substantially. Another way to define change points is by treating them as delimiters between different subsets of observed data. In this work, we adopt a version of the former definition - we are interested in the event that causes a subsequent observation to differ substantially from the previous ones.

In terms of methods, change point detection can be broadly divided into online and offline methods. Online methods \citep{adams2007bayesian} focus on detecting a change point in a stream of data, and are evaluated based on the ability to predict a change point \textit{before} it occurs (i.e.\ before the value changes substantially). On the other hand, offline \citep{smith1975bayesian, green1995reversible} methods by design work retrospectively on a sequence of datapoints, aiming at solving the task of segmentation. Offline methods incorporate bi-directional information to determine when a change point occurred (i.e.\ the data points before and after the change point), whilst online methods rely only on the observed information. In this work we focus exclusively on online change point detection, as
we would like to predict what causes a change 
before we observe the change it causes. Arguably, detecting a change of mind post-hoc should be a more trivial endeavour, as s model could learn cues such as agreement markers or solution proposals.

A different way to approach this would be from the point of view of survival analysis and reliability engineering \citep{Read2016HazardFA, Diamoutene2021ReliabilityAW, Nikulin2011ReliabilityAO}. Previous research relies on the concept of hazard function (also referred to as "time-to-failure") which is defined as the instantaneous risk of an event occurring at a point in time. The premise is that, as more time passes, the likelihood of an event occurring increases. Practically speaking, in engineering the hazard function captures the intuition that as more time passes since the last maintenance, the likelihood of a breakdown of apparatus increases. We hypothesise that we observe a similar phenomenon in conversations -- as the dialogue progresses, it is more likely for a participant to change their mind.




\section{Data}
\label{sec:data}
In this work, we are investigating what makes someone change their mind in group decision-making. In order to select a dataset to work on, we considered the following factors:
\begin{itemize}
    \item The dataset should contain group discussions.
    \item When engaging in conversation, the group should collaborate in order to reach a decision
    \item The conversation should have a quantifiable measure of success
\end{itemize}   
With these criteria in mind, there are two datasets that could be used - a corpus of people playing a photography geo-location game \citep{niculae2016conversational}, or a dataset of groups playing the Wason card selection task \citep{karadzhov2021delidata}. Unfortunately, the former is not publicly available, so in this work we focus on the latter. The dataset was introduced by our previous work \citep{karadzhov2021delidata}, and it aims at evaluating how people collaborate and engage in deliberation (henceforth we refer to it as DeliData). Each group is presented with 4 cards, each having a letter or a number on it. Then, the participants have to answer the question "Which card(s) should you tern to test the rule: \textbf{All cards with vowels on one side, have an even number on the other}" (see Figure~\ref{fig:example_conversation_intro}). The intuitive but wrong answer to the question is to turn the vowel and the even number, which is due to confirmation bias and is the most common answer given to the task. The correct answer is to turn over the vowel and the odd number. 

In our experimental setup in DeliData \citep{karadzhov2021delidata}, we formed groups of 2 to 5 participants, first asking each member of the group to solve the task on their own. Then all of the participants engaged in a discussion about the task, being able to submit intermediate and final solutions.  
Each participant, apart from payment for their participation, was offered a bonus for submitting the correct solution, i.e.\ selecting the correct cards. 
A conversation is successful if the final solutions submitted by each group member were on average more correct (in terms of number of cards selected correctly) than the initial ones before the conversation took place.
In DeliData \citep{karadzhov2021delidata}, we found that after discussing the solution, 64\% of the groups perform better at the Wason task, compared to their initial performances. Moreover, in 43.8\% of the groups who had at least one correct answer as their final solution, none of the participants had solved the task correctly by themselves, thus confirming the hypothesis that groups can perform better than even the most knowledgeable individual. 

Statistics of the DeliData corpus are presented in Table~\ref{tab:delidata_stats}. DeliData contains 500 dialogues, with an average length of 28 utterances per dialogue. Additionally, 50 of those dialogues are annotated with deliberation cues and other conversational phenomena, such as argument structure or when someone proposes a solution. In this work, we use the solution proposals as an indication of \textit{when} someone changes their minds, thus helping us identify \textit{what} made them change their mind. These annotations were carried out by 3 annotators in a controlled setting, with a high inter-annotator agreement (0.5-0.75 Cohen's kappa).

\begin{table}[ht!]
    \centering
    \begin{tabular}{l|r}
        \# of dialogues & 500  \\ \hline
        \# of utterances & 14003 \\ \hline
       AVG group size & 3.16 \\ \hline
     \# of dialogues with \\ intermediary submissions & 220 \\ \hline
      \# of intermediary and \\ final submissions & 1179 \\ \hline
       \# of annotated dialogues & 50 \\ \hline
       {\# of annotated change of mind} & 262 \\ \hline
    \end{tabular}
    \caption{DeliData statistics}
    \label{tab:delidata_stats}
\end{table}


\subsection{Gold data}
In order to evaluate what made someone change their mind, we take advantage of the 50 dialogues manually annotated by \citet{karadzhov2021delidata}. If an utterance contains a solution proposal that is different to the previously proposed solution by the same participant, it is considered an expression of a change of mind. 
Leveraging this annotation, our gold data is defined as follows: Given an utterance that expresses a change of mind, we select the last utterance made by a different person as the utterance that caused this change of mind. In Figure~\ref{fig:data_splits} (left), the 3rd utterance is an expression of a change of mind, annotated in DeliData. Therefore, the last utterance not said by participant U1, would be considered what caused the change of mind.

\subsection{Weakly supervised training set}
Given that the gold annotated data is limited, we devised a way to leverage the unannotated data as a weakly annotated training set. For the 450 unannotated dialogues, each participant had to submit at least 1 solo solution, and 1 final solution. In 220 of these dialogues, at least 1 user had submitted an intermediate submission, thus we consider these dialogues as our training data.

\begin{figure}
    \centering
    \includegraphics[scale=0.45]{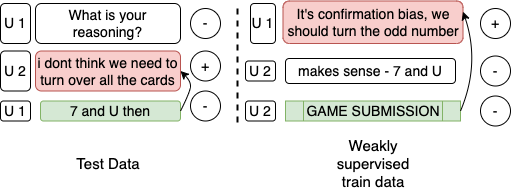}
    \caption{Test (left) and  weakly supervised training (right) data for what caused a change of mind. The circles on the right of each example show the annotation used in our experiments: + denotes an utterance that caused a change of mind}
    \label{fig:data_splits}
\end{figure}

Following the approach used for the gold data, we consider these weak annotations with a similar rule: for every submission expressing a different solution, we select the last utterance not made by the same user as the utterance that made them change their mind. In the example in Figure~\ref{fig:data_splits} (right), participant U2 made a submission, but because the last utterance before the submission was made by them, we mark the utterance by participant U1 as the one that made them change their mind.

As already mentioned, for this weakly supervised data we assume that every time a participant submits a new solution to the game, it can be attributed to the most recent utterance by a different participant.
While this is reasonable, the reverse is not true - we can't be sure that if someone changed their mind, they submitted a new solution. Hence there will be utterances that could have caused a change of mind, but are not annotated as such. Therefore in our training protocol we take into account this limitation by proposing the learning-to-rank training described in Section~\ref{ssec:rank}. 

\section{Models}
\label{sec:models}
\subsection{Language-agnostic models}
First, we consider modelling options that do not utilise the language directly, but rather rely on proxy signal to predict when a change point would occur. In particular, we investigate 3 variants for language-agnostic change point detection - Hazard Function, Sequence length probability, and Bayesian Online Changepoint Detection \citep{adams2007bayesian}.


\subsubsection{Hazard Function} 
\label{subsec:hazard}
 The hazard function encapsulates the intuition that an event is more likely to occur the more time passes.  It is defined as the probability of an event happening now, divided by the sum of probabilities of the event happening in the future. To calculate the hazard function we use Equation~\ref{eq:hazard_function}, where $Tcp$ denotes the number of time steps since the last change point, and $P(X_{Tcp} = CP)$ is the probability of change point occurring at time step $T$:
\begin{equation} \label{eq:hazard_function}
 H(Tcp) = \frac{P(X_{Tcp}=CP)} {\sum\limits_{t=Tcp}^{\infty} P(X_t=CP)}
\end{equation}
In the calculation of the hazard function, we consider only the distance from the last change point, thus disregarding information at what point of the conversation we are. Essentially, every time a change point occurs, the function starts over. For example, if a change point occurs in the 9th utterance of a conversation, the probability at the 10th utterance would be the same as in the first utterance.

\subsubsection{Sequence length Probability}
Recognising that it is important to model not only the information since the last change point, but to also consider information about how many utterances have been exchanged in the conversation as a whole, we propose an alternative model - sequence length probability. The assumption behind this method is that conversational turning points are more likely to occur at certain time steps in a conversation. For example, people may change their minds more at the end of a conversation rather than just after the first few utterances.
This approach estimates the likelihood of encountering a change point at a specific time step \textit{since the beginning} of the observed process. To model that, we calculate what is the chance for a change point occurring at time step $T$.

\subsubsection{Bayesian change point detection}
\citet{adams2007bayesian} 
proposed a Bayesian approach to modelling when a change point would occur. Their method performs a prediction based on two variables - time since the last change point (similarly to the hazard function) and an observed variable at each time step. In the case of the DeliData dataset, we extract the observed variable from a method we call \textbf{solution tracker}, which gives an estimate what is the group's performance at every utterance. The solution tracker keeps a record of the solution proposed by each participant and then averages their individual score to calculate the group performance. The solution tracker first records each of the participants' solo submission. After the group phase starts, every time a participant mentions one of the 4 cards or the words '\textit{all}' and '\textit{none}', the solution tracker recalculates participant's individual score, as well as the aggregated group performance. The solution tracker incorporates a fairly simplistic rule-based approach to tracking solutions, and is thus imperfect. Nevertheless, it is a reasonable measure to track as it is a proxy for team performance.

Following the approach introduced by \citet{adams2007bayesian}, we are interested in two probabilities - the growth probability, indicating that a change point will not occur in the next time step (Equation~\ref{eq:ocp_growth}), and the change point probability, showing that a change point would occur (Equation~\ref{eq:ocp_cp}). 
\begin{equation} \label{eq:ocp_growth}
\begin{split}
P(X_{T+1} \neq CP) = \\ P_{r}(T-1)\pi_{T}^{(r)}(1-H(r_{t-1}))
\end{split}
\end{equation}

\begin{equation} \label{eq:ocp_cp}
\begin{split}
P(X_{T+1}=CP) = \\ \sum_{r_{t-1}} P_{r}(t-1) \pi_{t}^{(r)} H(r_{t-1})
\end{split}
\end{equation}
These probabilities are computed using: 
\begin{itemize}
\item $P_{r}(t-1)$ -  run length estimation, which is the probability of the length of the run since the last change point, given the observed data and the current time step
\item $\pi_{t}^{(r)} = P(X_1..X_t)$  - predictive probability, i.e. how likely is to observe a specific sequence of values
\item $H(r_{t-1})$ - hazard function, as described in section~\ref{subsec:hazard} and equation~\ref{eq:hazard_function}
\end{itemize}

It is important to note that \citet{adams2007bayesian} consider the model parameter before and after the change point as independent of each other, thus any positional information is lost. 

\subsection{Text-based Models}
We recognise that the output of the solution tracker is unlikely to contain all the information needed to determine whether an utterance would cause someone to change their mind, hence we experimented with linguistic models to perform this prediction. We use a neural network, where the input is the last two utterances of a certain time step in a conversation, and the predicted output is whether or not we will encounter a change of mind in the next time step.  Henceforth we will refer to this model as the \textbf{linguistic model}.

\subsection{Learning to rank training}
\label{ssec:rank}
Given that changes of mind are often not stated by the participants, we presume that the annotation of the utterances that causes them would be incomplete, and we will be dealing with a lot of false negatives in the training. 
Thus, we propose to use learning to rank as follows: given a pair of inputs, one that is annotated as a cause of a change of mind and one that is not, we use the model to score the positive input higher than the negative one. In other words, even if both of the inputs are predicted as not causing a change of mind, we adjust the loss so that the positive sample should be ranked higher than the negative one. 
\par
Since the positive class is substantially less prevalent than the negative (most utterances do not change minds), we only need as many negative samples as there are positive ones to construct the positive-negative pairs. 
To do this, we devise the following algorithm:
\begin{itemize}[noitemsep,nolistsep]
    \item For each positive input (an utterance causing a change of mind) in a dialogue, we select a random negative input from the same dialogue.
    \item When selecting a random negative input, we consider those that are with a distance greater than 2 utterances from the nearest change point. This allows us to select safer negative inputs, as opposed to those that might carry a partial signal of the cause of change of mind. 
    \item For every training epoch we change the random seed for the selection of the negative sample while keeping the same positive samples. 
\end{itemize}

Using this algorithm, the positive samples are consistent throughout the training, while we vary the negative ones. 

\par

Having this training procedure, we consider \textbf{RankNet} loss \cite{burges2005learning_ranknet}, presented on Equation~\ref{eq:ranknet}, which is a modified logistic function on probabilities from \citet{baum1987supervised}. This loss provides a probabilistic ranking cost function, which relies only on the difference between the positive and the negative samples.
\begin{equation} \label{eq:ranknet}
    C(pos, neg) = 1 - \frac{e^{(pos - neg)}}{1 + e^{(pos - neg)}}
\end{equation}
The inspiration for this type of training was drawn from a different area in machine learning research - recommender systems.
There, a single user will interact with a limited number of items from the pool of available ones. For even fewer of those, the user would have provided positive feedback. Therefore there will be items that the user would like to see more of, but they have not provided positive feedback, hence having incomplete annotation. 
When detecting a cause for a change of mind, we observe similarly incomplete annotation - not every time someone changed their mind, they have expressed it in the conversation explicitly, and thus we cannot label which utterance would be the cause for that change of mind. However, similarly to recommender systems, positive feedback while rarer is a strong indication of when a change of mind occurs. Therefore, in this work, we propose to approach the task of predicting conversational turning points using a learning to rank training objective, rather than as standard classification.

\section{Evaluation}
\label{sec:eval}

\begin{figure}[ht!]
    \centering
    \includegraphics[scale=0.295]{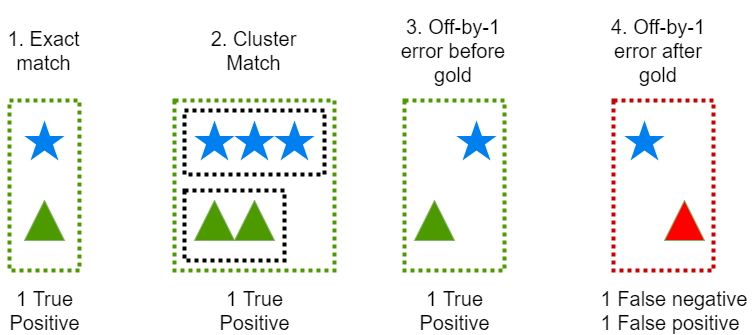}
    \caption{Four scenarios for change of mind evaluation. Blue stars denote the gold labels, while the triangles show the predicted values. With green borders and triangles, we show where the predicted and gold values match, and with red where we have an inaccuracy.}
    \label{fig:eval}
\end{figure}

In this work, we devise a novel way for evaluating what caused a change of mind. 
We propose 3 key properties that our evaluation method should exhibit (with corresponding examples on Figure~\ref{fig:eval}):

\begin{itemize}[noitemsep,nolistsep]
    \item The method should reward exact matches, i.e.\ when the gold and the predicted cause of change of mind align perfectly. (Scenario 1 on Figure~\ref{fig:eval}) 
    \item In cases where we observe a cluster of causes of changes of minds, we would like our method to (i) give full credit if we predict at least 1 of the gold utterances in the cluster, and (ii) if we predict all of them, to not "inflate" the score, giving full credit for each. An example of that is presented in Figure~\ref{fig:eval}, scenario 2 -- the gold and the predicted labels are clustered and aligned. Given that we have alignment between the two clusters, the method should count this as 1 true positive.
    \item In order to provide a more relaxed evaluation, our method should allow for small-margin errors. Given the length of each dialogue, we set the margin to 1. That said, we should count a true positive for off-by-one errors before the cause of change of mind (scenario 3), but the method should not allow for off-by-one errors after the gold label (scenario 4).
\end{itemize}

Given these desiderata, we consider how previous work evaluates change point detection methods. One approach \cite{killick2012optimal} is to evaluate such methods as a regular machine learning model - the predicted and gold events should match exactly, in order to count the change point as true positive. This would cover scenario 1 (exact match), will count 2 true positives and 1 false negative for scenario 2 (cluster match) and will count the off-by-one predictions as errors. Some approaches \citep{martin2004learning} for change point detection evaluation recognise that nearly predicting a change point is good enough in practice, thus allowing for off-by-one errors. In Figure~\ref{fig:eval}, both scenarios 3 and 4 are concerned with off-by-one errors, and previous work would categorise both of these as true positive, which would be incorrect (as we are not allowing off-by-one errors after the gold label).

Taking into consideration the desired properties of our evaluation method, as well as the limitations of previous research, in this work we use the following evaluation procedure:

\begin{enumerate}[noitemsep,nolistsep]
    \item We identify all clusters in the predicted and in the gold sequences, by grouping instances that are consecutive.
    \item We perform alignment to identify which clusters overlap. We consider 2 clusters aligned if they overlap by at least 1 element. 
    \item We identify all matches between the gold and predicted pairs. If we encounter a gold label or cluster of labels, we check the prediction at the current and the previous time steps. If there is a match, we consider this pair a true positive.
    \item After we iterate through the gold-prediction sequences, we mark every gold utterance that was not matched to a prediction as a false negative. Likewise, every predicted utterance that did not match a gold label, is considered a false positive.
\end{enumerate}

Given this training procedure, we are able to have a list of true positive, false positive and false negative cases for our test set, allowing us to calculate class  measures such as area under the precision-recall curve, and break-even precision-recall point. 

\section{Experimental Setup}
Using the gold and the weakly supervised sets introduced in section~\ref{sec:data}, we train all of our models with the following setup. All models are trained on the 220 dialogues from the weakly supervised set. The 50 gold annotated dialogues are split into test and validation sets, of 40 and 10 dialogues respectively. The validation sets are exclusively used for model selection of the text-based models. 

To train the linguistic and learning-to-rank models, we used a similar training setup. Both models embed the input using the BART embedding layer \citep{bart}.
Following, we added two fully-connected layers with size of 1024 and 0.3 dropout between each of the layers. Finally, we use a simple sigmoid function to perform the final classification. Both models are trained using a batch size of 32 using the Adadelta optimizer \citep{Zeiler2012ADADELTAAA} and were trained for 100 epochs, saving the iteration that has the best area under the precision-recall curve.

\section{Results}
\begin{table*}[ht!]
    \centering
    \scalebox{0.8}{
    \begin{tabular}{|l|l|l|l|}
        \hline
        \textbf{Model} & \textbf{Micro AUC} & \textbf{Macro AUC} & \textbf{Cutoff}  \\ \hline
        \textbf{Baseline}: All positive & 0.07 & 0.07 & N/A \\ \hline
        \textbf{Baseline}:  Bag of Words & 0.19 & 0.20 & 0.21 \\ \hline \hline
        \textbf{[1]} Hazard Function & 0.16 & 0.17 & 0.16 \\ \hline
        \textbf{[2]} Sequence Length & 0.17 & 0.17 & 0.20 \\ \hline
        \textbf{[3]} BOCP \cite{adams2007bayesian} & 0.18 & 0.21 & 0.22 \\ \hline
        \textbf{[4]} [2] + [3] & 0.21 & 0.23 & 0.26 \\ \hline
        \textbf{[5]} Linguistic Model & 0.20 & 0.20 & 0.23 \\ \hline
        \textbf{[6]} Linguistic Model + [4] & 0.22 & 0.22 & 0.24 \\ \hline
        \textbf{[7]} Linguistic Model (Learning to Rank) & 0.23 & \textbf{0.26} & 0.24 \\ \hline 
        \textbf{[8]} Linguistic Model (Learning to Rank) + [4] &\textbf{0.25} & \textbf{0.26} & \textbf{0.30} \\ \hline 

    \end{tabular}}
    \caption{Evaluation of different methods for detecting what causes a change of mind in group discussions}
    \label{tab:results}
\end{table*}

On Table~\ref{tab:results} we compare all of the models introduced in section~\ref{sec:models}, together with two baseline models. As a naïve baseline, we predict that every utterance in the conversation will lead to a change of mind. Also, we use off-the-shelf text classification methods, to provide a basic baseline for a linguistic model, namely a bag-of-words approach, paired with a Random Forest classifier \citep{ho1995random}. 
\par
We use 3 evaluation measures to compare the models - micro (utterance level) and macro (dialogue level) averaged area under the precision-recall curve, and the precision-recall cutoff point - the point where the precision and recall are equal. The reason to use these evaluation measures is three-fold. First, since change point detection typically deals with very imbalanced data, we need measures that are robust when the class of interest is under-represented. When dealing with heavily skewed data, \citet{davis2006relationship} argue that the area under the ROC curve gives an overly optimistic estimate of the performance, and thus area under the precision-recall curve is a more appropriate measure. Secondly, while evaluating our models, we noticed that different models have different precision and recall characteristics. For example, some of our models had very high precision, or very high recall, whilst producing comparable F-measures. In order to give a  fairer comparison of the overall model performance, we report the micro- and macro- average area under the precision-recall curve. Finally, while area-under-the-curve gives a good estimation of performance, it doesn't give a lot of intuition of how the model will perform in terms of precision and recall when used in a practical setting. Thus, we also report the precision-recall break-even point to show the relative predictive power of each model.

\par
In Table~\ref{tab:results} we show that all of the methods outperform the "all positive" baseline. That said, using the hazard function and the sequence length probability by themselves are the worst performing methods. Better performance is achieved by using a more sophisticated language-agnostic modelling, the Bayesian online change point detection (BOCP) (result 3). This approach takes into account the hazard function as well as a proxy for conversation performance, thus allowing for better modelling. 
While these approaches are reasonable, 
they are unable 
to capture language such as the arguments being made, which may cause lower performance. 
Interestingly, the bag-of-words model performs similarly to the significantly larger neural linguistic model which is trained on top of BART \cite{bart} (result 5).

The best performing stand-alone model is achieved by training the linguistic model in a learning-to-rank setup (result 7), achieving Micro and Macro averaged AUC of 0.23 and 0.26 respectively.

\par

Further, we experimented with combining language-independent and language-agnostic models. In the context of this paper, we incorporated a simplistic combination - if either of the combined models predicts a conversational turning point we consider this as a positive signal.  
Analysing the results, we observe that incorporating the sequence length and the BOCP \cite{adams2007bayesian} with all linguistic models can yield a substantial improvement. By combining the sequence length with the Neural Learning-to-Rank model, we improve the performance to 0.25 micro AUC and 0.30 P-R break-even point. 

\par

In summary, while the neural models provide good stand-alone performance, they don't capture all of the information required for a prediction of a conversational turning point. Namely, a substantial signal is carried by a positional information of where you are in a dialogue (captured by the sequence length probability), as well as patterns in how people discuss solutions (Bayesian online change-point detection).

\section{Qualitative Study}

In order to gain some understanding of how each of the methods works, we qualitatively evaluate models' predictions. Full conversation and model predictions are presented in appendix~\ref{app:example}. We use LIME \citep{ribeiro2016should} to find out which words are indicative for the positive predictions. We incorporated LIME's to explain the prediction of the positive (cause of change of mind) class. The way method works is by first randomly perturbing features from the input, and then by learning linear models on the neighbourhood data to explain the label of interest. Using this workflow, we extracted common words for each of the methods and for the rest of the section we present some of the findings.

If we consider a pair of utterances that contain group interaction in the form of user mention: ``\textit{utt1: <MENTION> any ideas ? utt2: but then again most people get this wrong then it cant be as easy as we think surely}''. Here both the bag-of-words baseline and the neural linguistic model classified the second utterance as a cause of change of mind. Interestingly, the models gave weight to different features. The bag-of-words identified words such as ``easy'', ``people'' and ``wrong'' as important, which are part of an argument. On the other hand, the neural linguistic model put by far the highest weight on the participant mention, which is not related to the task at hand, but rather to the group dynamics. This observation is also supported by previous research \citep{niculae2016conversational, woolley2010evidence}, which argues that group dynamics play important role in collaboration.

Looking into the cases where one of the models predicted a cause of change of mind one utterance before the cause (as we allow for off by 1 errors), we consider the following pair of utterances:  ``\textit{then yeah we d have to make sure two vowels or two even numbers appear <SEP> so i think you'd just need to turn over <CARD> and <CARD>}''. Here the neural learning-to-rank model, predicted a cause of change of mind, and some of the words with the highest weight were ``odd'', ``turn'', and ``need''. We hypothesise that the model learned to recognise argument markers as suggestive for a cause of future change of mind. Similarly, in the example ``<CARD> is not an even we know tat <SEP> that*'' the learning-to-rank model put higher weights on the words ``even'', <CARD> and ``know''.

Generally, the qualitative analysis shows that our best model (learning-to-rank) learnt to recognise argument cues as indicative of a conversational turning point. The model identified words that are related to the task such as card mentions or specific terms of the Wason card selection task. That said, this could be a drawback - it is unclear how such models would perform for a different task, where the vocabulary is substantially different.

\section{Conclusions}
In this work, we investigated methods for detecting the utterances that make someone change their mind, 
in the context of a recently introduced dataset containing group discussions of people collaborating to solve a task.
We demonstrate that the best performance is achieved by combining a text-based model with a language-agnostic ones (such as positional information). In future work, we want to leverage the proposed approach to develop a system that can generate utterances that cause a change of mind in order to enhance group decision-making.

\section*{Acknowledgements}
The authors would like to acknowledge the support of the Isaac Newton Trust and Cambridge University Press in creating the DeliData dataset \cite{karadzhov2021delidata}. 
Georgi Karadzhov is supported by EPSRC doctoral training scholarship. Tom Stafford and Andreas Vlachos are supported by the EPSRC grant no.\ EP/T023414/1: Opening Up Minds.

\bibliographystyle{acl_natbib}
\bibliography{anthology,custom}

\begin{thebibliography}{38}
\expandafter\ifx\csname natexlab\endcsname\relax\def\natexlab#1{#1}\fi

\bibitem[{Adams and MacKay(2007)}]{adams2007bayesian}
Ryan~Prescott Adams and David~JC MacKay. 2007.
\newblock Bayesian online changepoint detection.
\newblock \emph{stat}, 1050:19.

\bibitem[{Baum and Wilczek(1987)}]{baum1987supervised}
Eric Baum and Frank Wilczek. 1987.
\newblock Supervised learning of probability distributions by neural networks.
\newblock In \emph{Neural information processing systems}.

\bibitem[{Burg and Williams(2020)}]{burg2020evaluation}
GJJ Burg and CKI Williams. 2020.
\newblock An evaluation of change point detection algorithms.
\newblock \emph{arXiv preprint arXiv:2003.06222}.

\bibitem[{Burges et~al.(2005)Burges, Shaked, Renshaw, Lazier, Deeds, Hamilton,
  and Hullender}]{burges2005learning_ranknet}
Chris Burges, Tal Shaked, Erin Renshaw, Ari Lazier, Matt Deeds, Nicole
  Hamilton, and Greg Hullender. 2005.
\newblock Learning to rank using gradient descent.
\newblock In \emph{Proceedings of the 22nd international conference on Machine
  learning}, pages 89--96.

\bibitem[{Chen and Gupta(1997)}]{chen1997testing}
Jie Chen and Arjun~K Gupta. 1997.
\newblock Testing and locating variance changepoints with application to stock
  prices.
\newblock \emph{Journal of the American Statistical association},
  92(438):739--747.

\bibitem[{Concannon et~al.(2015)Concannon, Healey, and
  Purver}]{concannon2015shifting}
Shauna Concannon, Patrick~GT Healey, and Matthew Purver. 2015.
\newblock Shifting opinions: An experiment on agreement and disagreement in
  dialogue.
\newblock \emph{SEMDIAL 2015 goDIAL}, page~15.

\bibitem[{Davis and Goadrich(2006)}]{davis2006relationship}
Jesse Davis and Mark Goadrich. 2006.
\newblock The relationship between precision-recall and roc curves.
\newblock In \emph{Proceedings of the 23rd international conference on Machine
  learning}, pages 233--240.

\bibitem[{De~Kock and Vlachos(2021)}]{de2021beg}
Christine De~Kock and Andreas Vlachos. 2021.
\newblock I beg to differ: A study of constructive disagreement in online
  conversations.
\newblock In \emph{Proceedings of the 16th Conference of the European Chapter
  of the Association for Computational Linguistics: Main Volume}, pages
  2017--2027.

\bibitem[{Diamoutene et~al.(2021)Diamoutene, Noureddine, Kamsu-Foguem, and
  Barro}]{Diamoutene2021ReliabilityAW}
Abdoulaye Diamoutene, Farid Noureddine, Bernard Kamsu-Foguem, and Diakarya
  Barro. 2021.
\newblock Reliability analysis with proportional hazard model in aeronautics.
\newblock \emph{International Journal of Aeronautical and Space Sciences},
  pages 1--13.

\bibitem[{Green(1995)}]{green1995reversible}
Peter~J Green. 1995.
\newblock Reversible jump markov chain monte carlo computation and bayesian
  model determination.
\newblock \emph{Biometrika}, 82(4):711--732.

\bibitem[{Hallsson and Kappel(2020)}]{hallsson2020disagreement}
Bj{\o}rn~G Hallsson and Klemens Kappel. 2020.
\newblock Disagreement and the division of epistemic labor.
\newblock \emph{Synthese}, 197(7):2823--2847.

\bibitem[{Hensman et~al.(2013)Hensman, Lawrence, and
  Rattray}]{hensman2013hierarchical}
James Hensman, Neil~D Lawrence, and Magnus Rattray. 2013.
\newblock Hierarchical bayesian modelling of gene expression time series across
  irregularly sampled replicates and clusters.
\newblock \emph{BMC bioinformatics}, 14(1):1--12.

\bibitem[{Hidey et~al.(2017)Hidey, Musi, Hwang, Muresan, and
  McKeown}]{hidey-etal-2017-analyzing}
Christopher Hidey, Elena Musi, Alyssa Hwang, Smaranda Muresan, and Kathy
  McKeown. 2017.
\newblock \href {https://doi.org/10.18653/v1/W17-5102} {Analyzing the semantic
  types of claims and premises in an online persuasive forum}.
\newblock In \emph{Proceedings of the 4th Workshop on Argument Mining}, pages
  11--21, Copenhagen, Denmark. Association for Computational Linguistics.

\bibitem[{Ho(1995)}]{ho1995random}
Tin~Kam Ho. 1995.
\newblock Random decision forests.
\newblock In \emph{Proceedings of 3rd international conference on document
  analysis and recognition}, volume~1, pages 278--282. IEEE.

\bibitem[{Karadzhov et~al.(2021)Karadzhov, Stafford, and
  Vlachos}]{karadzhov2021delidata}
Georgi Karadzhov, Tom Stafford, and Andreas Vlachos. 2021.
\newblock Delidata: A dataset for deliberation in multi-party problem solving.
\newblock \emph{arXiv preprint arXiv:2108.05271}.

\bibitem[{Khapalova et~al.(2018)Khapalova, Jandhyala, Fotopoulos, and
  Overland}]{khapalova2018assessing}
Elena~A Khapalova, Venkata~K Jandhyala, Stergios~B Fotopoulos, and James~E
  Overland. 2018.
\newblock Assessing change-points in surface air temperature over alaska.
\newblock \emph{Frontiers in Environmental Science}, page 121.

\bibitem[{Killick et~al.(2012)Killick, Fearnhead, and
  Eckley}]{killick2012optimal}
Rebecca Killick, Paul Fearnhead, and Idris~A Eckley. 2012.
\newblock Optimal detection of changepoints with a linear computational cost.
\newblock \emph{Journal of the American Statistical Association},
  107(500):1590--1598.

\bibitem[{Lai(1995)}]{lai1995sequential}
Tze~Leung Lai. 1995.
\newblock Sequential changepoint detection in quality control and dynamical
  systems.
\newblock \emph{Journal of the Royal Statistical Society: Series B
  (Methodological)}, 57(4):613--644.

\bibitem[{Lewis et~al.(2019)Lewis, Liu, Goyal, Ghazvininejad, Mohamed, Levy,
  Stoyanov, and Zettlemoyer}]{bart}
Mike Lewis, Yinhan Liu, Naman Goyal, Marjan Ghazvininejad, Abdelrahman Mohamed,
  Omer Levy, Veselin Stoyanov, and Luke Zettlemoyer. 2019.
\newblock \href {http://arxiv.org/abs/1910.13461} {{BART:} denoising
  sequence-to-sequence pre-training for natural language generation,
  translation, and comprehension}.
\newblock \emph{CoRR}, abs/1910.13461.

\bibitem[{Maciejovsky et~al.(2013)Maciejovsky, Sutter, Budescu, and
  Bernau}]{maciejovsky2013teams}
Boris Maciejovsky, Matthias Sutter, David~V Budescu, and Patrick Bernau. 2013.
\newblock Teams make you smarter: How exposure to teams improves individual
  decisions in probability and reasoning tasks.
\newblock \emph{Management Science}, 59(6):1255--1270.

\bibitem[{Martin et~al.(2004)Martin, Fowlkes, and Malik}]{martin2004learning}
David~R Martin, Charless~C Fowlkes, and Jitendra Malik. 2004.
\newblock Learning to detect natural image boundaries using local brightness,
  color, and texture cues.
\newblock \emph{IEEE transactions on pattern analysis and machine
  intelligence}, 26(5):530--549.

\bibitem[{Mercier and Sperber(2011)}]{mercier2011humans}
Hugo Mercier and Dan Sperber. 2011.
\newblock Why do humans reason? arguments for an argumentative theory.
\newblock \emph{Behavioral and brain sciences}, 34(2):57--74.

\bibitem[{Navajas et~al.(2018)Navajas, Niella, Garbulsky, Bahrami, and
  Sigman}]{navajas2018aggregated}
Joaquin Navajas, Tamara Niella, Gerry Garbulsky, Bahador Bahrami, and Mariano
  Sigman. 2018.
\newblock Aggregated knowledge from a small number of debates outperforms the
  wisdom of large crowds.
\newblock \emph{Nature Human Behaviour}, 2(2):126--132.

\bibitem[{Niculae and
  Danescu-Niculescu-Mizil(2016)}]{niculae2016conversational}
Vlad Niculae and Cristian Danescu-Niculescu-Mizil. 2016.
\newblock Conversational markers of constructive discussions.
\newblock In \emph{Proceedings of NAACL-HLT}, pages 568--578.

\bibitem[{Nikulin et~al.(2011)Nikulin, Saaidia, and
  Tahir}]{Nikulin2011ReliabilityAO}
Mikhail Nikulin, Noureddine Saaidia, and Ramzan Tahir. 2011.
\newblock Reliability analysis of redundant systems by simulation for data with
  unimodal hazard rate functions.
\newblock \emph{Journal | MESA}, 2:277--286.

\bibitem[{Oh and Han(2001)}]{oh2001intelligent}
Kyong~Joo Oh and Ingoo Han. 2001.
\newblock An intelligent clustering forecasting system based on change-point
  detection and artificial neural networks: Application to financial economics.
\newblock In \emph{Proceedings of the 34th Annual Hawaii International
  Conference on System Sciences}, pages 8--pp. IEEE.

\bibitem[{Page(1954)}]{page1954continuous}
Ewan~S Page. 1954.
\newblock Continuous inspection schemes.
\newblock \emph{Biometrika}, 41(1/2):100--115.

\bibitem[{Read and Vogel(2016)}]{Read2016HazardFA}
Laura~K. Read and Richard~M. Vogel. 2016.
\newblock Hazard function analysis for flood planning under nonstationarity.
\newblock \emph{Water Resources Research}, 52:4116 -- 4131.

\bibitem[{Reeves et~al.(2007)Reeves, Chen, Wang, Lund, and
  Lu}]{reeves2007review}
Jaxk Reeves, Jien Chen, Xiaolan~L Wang, Robert Lund, and Qi~Qi Lu. 2007.
\newblock A review and comparison of changepoint detection techniques for
  climate data.
\newblock \emph{Journal of applied meteorology and climatology},
  46(6):900--915.

\bibitem[{Ribeiro et~al.(2016)Ribeiro, Singh, and Guestrin}]{ribeiro2016should}
Marco~Tulio Ribeiro, Sameer Singh, and Carlos Guestrin. 2016.
\newblock " why should i trust you?" explaining the predictions of any
  classifier.
\newblock In \emph{Proceedings of the 22nd ACM SIGKDD international conference
  on knowledge discovery and data mining}, pages 1135--1144.

\bibitem[{Smith(1975)}]{smith1975bayesian}
Adrian~FM Smith. 1975.
\newblock A bayesian approach to inference about a change-point in a sequence
  of random variables.
\newblock \emph{Biometrika}, 62(2):407--416.

\bibitem[{Truong et~al.(2020)Truong, Oudre, and Vayatis}]{truong2020selective}
Charles Truong, Laurent Oudre, and Nicolas Vayatis. 2020.
\newblock Selective review of offline change point detection methods.
\newblock \emph{Signal Processing}, 167:107299.

\bibitem[{Turner et~al.(2013)Turner, Bottone, and Stanek}]{turner2013online}
Ryan~D Turner, Steven Bottone, and Clay~J Stanek. 2013.
\newblock Online variational approximations to non-exponential family change
  point models: with application to radar tracking.
\newblock \emph{Advances in Neural Information Processing Systems}, 26.

\bibitem[{Wang et~al.(2011)Wang, Wu, Ji, Wang, and
  Liang}]{wang2011non_cp_genes}
Yao Wang, Chunguo Wu, Zhaohua Ji, Binghong Wang, and Yanchun Liang. 2011.
\newblock Non-parametric change-point method for differential gene expression
  detection.
\newblock \emph{PloS one}, 6(5):e20060.

\bibitem[{Wason(1968)}]{wason1968reasoning}
Peter~C Wason. 1968.
\newblock Reasoning about a rule.
\newblock \emph{Quarterly journal of experimental psychology}, 20(3):273--281.

\bibitem[{Woolley et~al.(2010)Woolley, Chabris, Pentland, Hashmi, and
  Malone}]{woolley2010evidence}
Anita~Williams Woolley, Christopher~F Chabris, Alex Pentland, Nada Hashmi, and
  Thomas~W Malone. 2010.
\newblock Evidence for a collective intelligence factor in the performance of
  human groups.
\newblock \emph{science}, 330(6004):686--688.

\bibitem[{Zeiler(2012)}]{Zeiler2012ADADELTAAA}
Matthew~D. Zeiler. 2012.
\newblock Adadelta: An adaptive learning rate method.
\newblock \emph{ArXiv}, abs/1212.5701.

\bibitem[{Zeng et~al.(2020)Zeng, Li, He, Gao, Lyu, and
  King}]{zeng2020whatchanged}
Jichuan Zeng, Jing Li, Yulan He, Cuiyun Gao, Michael Lyu, and Irwin King. 2020.
\newblock What changed your mind: The roles of dynamic topics and discourse in
  argumentation process.
\newblock In \emph{Proceedings of The Web Conference 2020}, pages 1502--1513.

\end{thebibliography}

\clearpage
\appendix
\section{Dialogue example and model predictions}
\label{app:example}
\noindent

    \centering
    \scalebox{0.8}{
    \begin{tabular}{|l|c|c|c|c|c|c|c|}
    \hline 
 Utterance & Gold & OCP & Hazard & SeqLen & Ling & BoW & L2R \\ \hline 
What did you guys say was the answer ? & 1 & 0 & 0 & 0 & 1 & 0 & 0 \\ \hline 
<CARD> is not an even we know tat & 0 & 0 & 0 & 0 & 0 & 0 & 0 \\ \hline 
that * & 0 & 0 & 0 & 0 & 0 & 0 & 1 \\ \hline 
i put <CARD> and <CARD> , you ? & 1 & 0 & 0 & 0 & 0 & 0 & 0 \\ \hline 
<CARD> , <CARD> and <CARD> & 0 & 0 & 0 & 0 & 0 & 1 & 0 \\ \hline 
Why did you think it was n't <CARD> ? & 1 & 1 & 0 & 0 & 0 & 0 & 0 \\ \hline 
i chose all 4 cards so clearly mine was n't the one & 0 & 0 & 0 & 0 & 0 & 0 & 0 \\ \hline 
Urm i m thinking & 0 & 0 & 0 & 0 & 0 & 0 & 0 \\ \hline 
It might be right , we need to discuss & 0 & 0 & 0 & 0 & 0 & 0 & 0 \\ \hline 
what do they exactly mean by turn & 0 & 0 & 0 & 0 & 0 & 0 & 0 \\ \hline 
turn over ? & 0 & 0 & 0 & 0 & 0 & 0 & 0 \\ \hline 
yeah & 0 & 0 & 0 & 0 & 0 & 0 & 0 \\ \hline 
I assumed so & 0 & 0 & 0 & 0 & 0 & 0 & 0 \\ \hline 
\multicolumn{1}{|p{8cm}|}{So what reasoning did you guys use for the cards you picked} & 0 & 0 & 0 & 0 & 0 & 0 & 0 \\ \hline 
\multicolumn{1}{|p{8cm}|}{they said most peope get this wrong so i m just wondering if they are trying to be cheeky by rotating them} & 0 & 0 & 0 & 0 & 0 & 0 & 0 \\ \hline 
why did you guys put your answers down ? & 1 & 0 & 0 & 0 & 0 & 0 & 1 \\ \hline 
\multicolumn{1}{|p{8cm}|}{ No , I think it means turning them over like onto the other side} & 0 & 0 & 0 & 0 & 0 & 0 & 0 \\ \hline 
\multicolumn{1}{|p{8cm}|}{Okay , I thought we need <CARD> because we need to see if there is a vowel on the other side} & 0 & 0 & 0 & 0 & 0 & 0 & 1 \\ \hline 
The same for <CARD> but the other way around & 0 & 0 & 0 & 0 & 0 & 0 & 0 \\ \hline 
yeah makes sense & 1 & 0 & 0 & 0 & 0 & 0 & 0 \\ \hline 
\multicolumn{1}{|p{8cm}|}{And <CARD> to see if the ' All ' section of the statement is correct} & 0 & 0 & 0 & 0 & 1 & 0 & 1 \\ \hline 
<MENTION> any ideas ? & 0 & 0 & 0 & 0 & 0 & 0 & 0 \\ \hline 
\multicolumn{1}{|p{8cm}|}{but then again most people get this wrong then it cant be as easy as we think surely} & 1 & 0 & 0 & 0 & 1 & 1 & 0 \\ \hline 
Probably not & 0 & 0 & 0 & 0 & 0 & 1 & 0 \\ \hline 
So do we think we should flip <CARD> ? & 1 & 0 & 0 & 0 & 0 & 0 & 0 \\ \hline 
\multicolumn{1}{|p{8cm}|}{then yeah we d have to make sure two vowels or two even numbers appear} & 0 & 0 & 0 & 0 & 0 & 0 & 1 \\ \hline 
\multicolumn{1}{|p{8cm}|}{so i think you d just need to turn over <CARD> and <CARD>} & 0 & 0 & 0 & 1 & 0 & 0 & 1 \\ \hline 
Why not <CARD> ? & 1 & 0 & 0 & 0 & 0 & 0 & 0 \\ \hline 
yeah and <CARD> like you said & 0 & 0 & 0 & 0 & 0 & 1 & 0 \\ \hline 
i m happy with that if you guys are & 0 & 0 & 0 & 1 & 1 & 0 & 0 \\ \hline 
I am & 0 & 0 & 0 & 0 & 1 & 1 & 0 \\ \hline 
yeah m happy with that & 0 & 0 & 0 & 0 & 0 & 0 & 0 \\ \hline 
i m * & 0 & 0 & 0 & 0 & 0 & 0 & 0 \\ \hline 
So <CARD> , <CARD> and <CARD> ? & 1 & 0 & 0 & 0 & 1 & 0 & 1 \\ \hline 
<CARD> , <CARD> \& <CARD> & 0 & 0 & 0 & 0 & 0 & 1 & 1 \\ \hline 
    \end{tabular}}

    \label{tab:sample_conversation}


\end{document}